\documentclass[letterpaper]{article} % DO NOT CHANGE THIS
\usepackage{aaai2026}  % DO NOT CHANGE THIS
\usepackage{times}  % DO NOT CHANGE THIS
\usepackage{helvet}  % DO NOT CHANGE THIS
\usepackage{courier}  % DO NOT CHANGE THIS
\usepackage[hyphens]{url}  % DO NOT CHANGE THIS
\usepackage{graphicx} % DO NOT CHANGE THIS
\usepackage{natbib}  % DO NOT CHANGE THIS AND DO NOT ADD ANY OPTIONS TO IT
\usepackage{caption} % DO NOT CHANGE THIS AND DO NOT ADD ANY OPTIONS TO IT
\usepackage{algorithm}
\usepackage{algorithmic}
\usepackage{multirow} 
\usepackage{amsmath} 
\usepackage{xcolor}

\usepackage{amsfonts}
\usepackage{bm}
\usepackage{newfloat}
\usepackage{listings}
\DeclareCaptionStyle{ruled}{labelfont=normalfont,labelsep=colon,strut=off} % DO NOT CHANGE THIS
\floatstyle{ruled}
\newfloat{listing}{tb}{lst}{}
\floatname{listing}{Listing}
\nocopyright 
\title{Rethinking Long-tailed Dataset Distillation: A Uni-Level Framework with Unbiased Recovery and Relabeling}
\author{Xiao Cui\textsuperscript{\rm 1,2}, Yulei Qin\textsuperscript{\rm 3}, Xinyue Li\textsuperscript{\rm 1,2}, Wengang Zhou\textsuperscript{\rm 1}\thanks{Corresponding authors: Wengang Zhou, Hongsheng Li and Houqiang Li.}, Hongsheng Li\textsuperscript{\rm 2}\textsuperscript{\textasteriskcentered}, Houqiang Li\textsuperscript{\rm 1}\textsuperscript{\textasteriskcentered}}  
\affiliations {
    \textsuperscript{\rm 1}University of Science and Technology of China \\
    \textsuperscript{\rm 2}The Chinese University of Hong Kong, MMLab\\
    \textsuperscript{\rm 3}Independent Researcher   \\
    \{cuixiao2001,lrel7\}@mail.ustc.edu.cn, yuleichin@126.com,  \{zhwg,lihq\}@ustc.edu.cn, hsli@ee.cuhk.edu.hk
}
\usepackage{bibentry}
\begin{document}

\maketitle
% \makeatletter{\renewcommand*{\@makefnmark}{}
% \footnotetext{Corresponding authors: Wengang Zhou, Hongsheng Li and Houqiang Li.}\makeatother}

\begin{abstract}
%好像有点太长了
% Dataset distillation aims to synthesize a small yet informative synthetic dataset that retains the core characteristics of a much larger one, enabling efficient training with significantly reduced data. 
Dataset distillation creates a small distilled set that enables efficient training by capturing key information from the full dataset.
% While existing DD methods perform well on balanced datasets, they %fail to generalize under long-tailed distributions, 
% they struggle under long-tailed distributions, where the imbalanced class frequencies introduce severe biases into the statistical estimation across instances such as batch normalization (BN) statistics and thereafter deviates the learned representations.
% in both the learned representations and the estimation of BN statistics.
%statistical estimates.
% In this paper, we rethink long-tailed dataset distillation through the lens of representation alignment, with the goal of improving distillation accuracy by mitigating statistic biases inherent in models.
% by mitigating model and statistic bias.
While existing dataset distillation 
methods perform well on balanced datasets, they struggle under long-tailed distributions, where imbalanced class frequencies induce biased model representations and corrupt statistical estimates such as Batch Normalization (BN) statistics. 
% In this paper, we rethink long-tailed dataset distillation through the lens of statistical alignment, aiming to improve dataset distillation performance by jointly mitigating model bias and restoring fair statistical supervision.
In this paper, we rethink long-tailed dataset distillation by revisiting the limitations of trajectory-based methods, and instead adopt the statistical alignment perspective to jointly 
mitigate model bias and restore fair supervision.
% We leverage a uni-level optimization framework and address these challenges through three key components:
To this end, we introduce three dedicated components that enable unbiased recovery of distilled images and soft relabeling: (1) enhancing expert models (an observer model for recovery and a teacher model for relabeling) to enable reliable statistics estimation and soft-label generation;
% enable reliable statistics estimation
% and to improve soft labels;
%improve feature alignment and soft label informativeness, respectively;
(2) recalibrating BN statistics via a full forward pass %on the training set 
with dynamically adjusted momentum to reduce representation skew;
%(3) initializing a class-balanced synthetic dataset via a round-robin, confidence-aware selection strategy tailored to long-tailed settings. 
(3) initializing synthetic images by incrementally selecting high-confidence and diverse augmentations via a multi-round mechanism that promotes coverage and diversity.
%incrementally selecting highly-confident, non-redundant augmentations,
% ensuring
%with a multi-round mechanism to ensure diversity and representativeness.
%controlling the contributions of each real image
% contributes in a controlled and 
% in a class-balanced manner.
% Extensive experiments across four long-tailed benchmarks
% demonstrate consistent outperformance over state-of-the-art approaches under varying imbalance factors.
Extensive experiments on four long-tailed benchmarks show consistent improvements over state-of-the-art methods across %a range of %imbalance factors.
varying degrees of class imbalance.
% demonstrate that our method consistently outperforms state-of-the-art approaches across various imbalance factors. 
Notably, our approach improves top-1 accuracy by 15.6\% on CIFAR-100-LT and 11.8\% on Tiny-ImageNet-LT under IPC=10 and IF=10. %Code is available at \url{https://anonymous.4open.science/r/URR}.
\end{abstract}


\section{Introduction}

Dataset distillation (DD) is the process of synthesizing a significantly smaller yet representative dataset that retains the essential characteristics of an original, larger dataset~\cite{dd_begin,dd_comprehensive_review,survey}. By drastically reducing data volume, DD facilitates efficient model training and substantially reduces computational costs, rendering it particularly valuable for resource-constrained scenarios~\cite{Cui2025StreetSurfGS,federate2}. Beyond reducing computational burden, DD also offers a compact lens for studying how data distribution affects model learning~\cite{insight2,insight}.

\begin{figure}
    \centering
\includegraphics[width=0.97\linewidth]{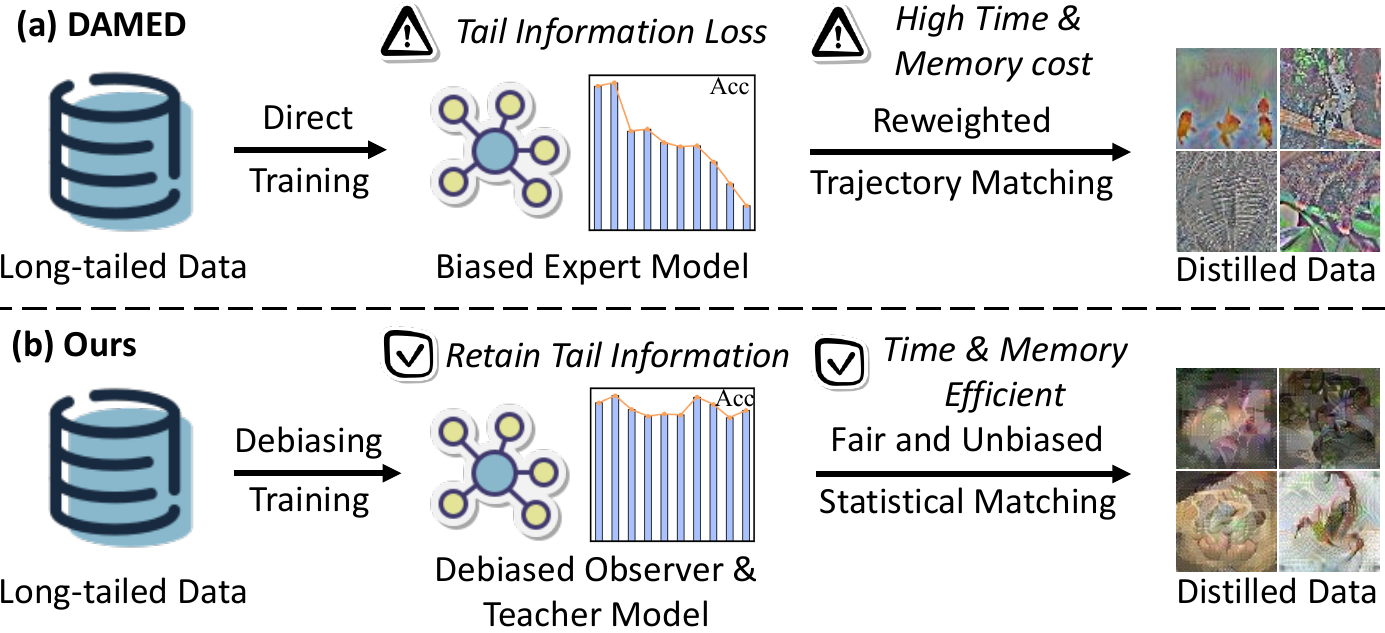}
    \caption{Comparison of DAMED and our method.
(a) DAMED directly trains a biased expert on long-tailed data and applies reweighted trajectory matching, leading to tail information loss and high computational cost.
(b) Our method debiases both observer and teacher models and performs unbiased statistical matching, effectively retaining tail knowledge with shorter time and lower memory cost.}
    \label{figure1}
\end{figure}

Long-tailed dataset distillation~\cite{damed} specifically addresses scenarios characterized by class imbalance, where a small subset of head classes contain abundant samples, whereas the remaining tail classes are sparsely represented. Such imbalance is ubiquitous in real-world applications as the acquisition of sufficient samples for rare categories is costly or infeasible. 
Most existing DD approaches~\cite{dream,datm,edc} perform well on balanced datasets, but struggle under class-imbalanced conditions. Their assumption of uniform data density leads to head-class dominance in the synthetic set and poor representation of minority classes, ultimately degrading performance in long-tailed scenarios.
% Most existing DD approaches~\cite{mttnew,ncf,delt} implicitly assume balanced class distributions and uniform data density,
% which inherently biases distilled datasets toward head classes. This leads to inadequate synthetic representation of minority classes and diminished performance when deployed on imbalanced datasets.

% Despite the clear significance,
Very few studies have explicitly tackled the limitations of conventional DD methods under long-tailed distributions, mainly because the widely used benchmarks typically feature balanced class structures~\cite{cifar,tiny-imagenet,deng2009imagenet}.
To our knowledge, DAMED~\cite{damed} is the only recent work that explicitly addresses this issue. 
% It introduces logit adjustments based on class frequencies to guide the student model toward expert-aligned gradient behavior, even when trained on balanced synthetic data. 
% It modifies the softmax computation by incorporating class-frequency-aware offsets, effectively adjusting the gradient magnitudes per class. This encourages the student model to mimic the expert's per-class gradient distribution, even when trained on balanced synthetic data.
It simulates long-tailed training dynamics by injecting class-frequency-aware offsets into the softmax layer, thereby %inducing head-class-dominated gradient behavior on balanced synthetic data.
inducing gradient behaviors that mimic those observed under imbalanced training.
However, as illustrated in Fig.~\ref{figure1}, DAMED still presents significant limitations.
% First, it trains the representation expert on long-tailed data without debiasing, resulting in biased features that under-represent tail classes.
1) \textbf{Under-representation of tail classes}.
% First, 
It relies on a feature extraction expert trained on long-tailed data but without debiasing, causing tail-class representations to be poorly preserved in the distilled dataset.
% Secondly, 
2) \textbf{Unintentional trade-off in trajectory matching}.
%Its gradient simulation also adversely affects mid-frequency classes, leading to the compromised performance across the distribution.
Mid-frequency classes receive unstable or insufficient gradient feedback, leading to compromised performance across the distribution.
% Moreover,
3) \textbf{Heavy computation overhead}.
Its bi-level trajectory-based optimization suffers from computational inefficiency and excessive GPU memory usage, severely restricting practical applicability~\cite{sre2l}.

% To comprehensively resolve these shortcomings, we propose a novel uni-level optimization framework that is both computationally efficient and explicitly designed to counteract biases stemming from class imbalance. In contrast to prior methods that neglect inherent biases within learned representations, we introduce a clear, principled problem formulation centered on statistical alignment. Our high-level approach systematically ensures unbiased, robust representations from both observer (for recovery) and teacher models (for relabeling), complemented by precise extraction and utilization of Batch Normalization (BN) statistics derived from the full long-tailed dataset.

% In contrast to prior methods that neglect inherent biases and overlook the semantic under-representation of tail classes, we introduce a clear, principled problem formulation centered on
% representation quality and
% fairness.
% Our high-level approach first explicitly debiases both the observer (for recovery) and the teacher models (for relabeling).
% We stem from the unbiased estimation of statistics,
% namely the most commonly used
% Batch Normalization (BN) parameters,
% and supervise the fine-grained representation learning across classes.
% Instances from under-represented classes are emphasized via their improved contributions in the subsequent global alignment, ensuring
% consistency across the representation space.
To comprehensively resolve these shortcomings, we propose a novel uni-level optimization framework that explicitly counteracts biases stemming from class imbalance in a cost-efficient manner.
% with computational efficiency.
% This is especially critical because effective debiasing strategies often cause the expert’s training trajectory to deviate significantly from standard SGD, rendering trajectory matching unstable or even infeasible.
This formulation is critical because effective debiasing strategies often cause the expert’s training trajectory to deviate significantly from that induced by standard training, rendering trajectory matching unstable and challenging.
%Our framework addresses debiasing primarily through unbiased statistical alignment. Its effectiveness depends on two key factors: the initialization process, which governs the diversity and coverage of synthetic samples; and the extraction of BN statistics, which are easily skewed under imbalanced training. In parallel, we incorporate semantic supervision based on soft labels from a fixed teacher model, whose predictions often exhibit head-class bias that must be addressed. This auxiliary signal complements the statistical alignment by correcting label-level supervision.
Our framework considers debiasing through two complementary components: unbiased synthetic image recovery and unbiased soft relabeling. 
% The effectiveness of recovery under long-tailed settings relies on two critical factors: the initialization process, which governs the diversity and representativeness of synthetic data; and the extraction of BN statistics, which enables precise statistical alignment during the recovery process.
% To enable effective recovery in long-tailed settings, we first debias the observer model to serve as a reliable feature carrier and then extract fair BN statistics from it for accurate statistical alignment. We also adopt a diverse and representative initialization of synthetic data to strengthen recovery.
% Meanwhile, unbiased soft relabeling provides effective semantic supervision through soft labels generated by another fixed, well-trained expert model.
To achieve effective recovery in long-tailed settings, our framework ensures a diverse and representative synthetic image initialization, and leverages a debiased expert model (observer model) to perform a fair extraction of BN statistics for precise alignment. 
Meanwhile, unbiased soft relabeling provides effective semantic supervision through soft labels generated by another well-trained, debiased expert model (teacher model).
To implement this design, we introduce three tailored strategies to mitigate model bias, statistical unfairness, and suboptimal initialization under long-tailed distributions.
First, we propose a \textbf{mixture consistency} loss and a \textbf{class-wise debias} loss to regularize both the observer and teacher models.
The former ensures robust feature learning against multiple data augmentations and the latter adopts dynamic weighting to rebalance class-wise supervision.
% we regularize both the observer and teacher models with a mixture consistency loss for robust feature learning, and a dynamically-weighted rebalancing strategy to correct class imbalance. These two objectives co-address both representational and supervisory bias.
Second, we recalibrate the estimation of BN statistics to address the challenges of class imbalance and temporal dependency.
% BN statistics to improve estimation accuracy. 
% After freezing
We freeze the observer model and perform an efficient forward-only pass over the entire training set.
During this process,
our dynamically adjusted momentum ensures equal contribution from all samples within each class, eliminating \textbf{intra-class bias}.
% , particularly in tail classes.
We then average the per-class statistics to obtain a globally balanced estimate, removing \textbf{inter-class bias}.
% Second, we obtain more accurate BN statistics from the imbalanced dataset. After freezing the observer model, we perform forward inference across the entire training set with dynamically adjusted momentum, iteratively recalibrating BN parameters. This process ensures that each image within the same class contributes equally to the BN statistics, thereby mitigating class-wise representation imbalance. We further average the class-specific BN statistics to ensure neutrality across classes, mitigating the inherent distributional skew and yielding unbiased distilled representations.
% Third, we propose a confidence-aware initialization strategy for distilled images, where each image is augmented multiple times and high-confidence crops are incrementally selected in a round-robin fashion--ensuring each image contributes at most one crop per round. This incremental selection continues until a class-dependent budget is met. To support consistent memory layout across imbalanced classes, zero-initialized placeholders are used solely as structural padding.
Third, we introduce a confidence-aware, class-independent strategy for \textbf{synthesis initialization}.
% For each real image,
% multiple candidate variants are derived round-by-round and selected according to their labeled confidence of the teacher.
% To ensure sample-level diversity, each image contributes at most once per round,
% and additional augments are selected only if necessary.
% For each real image, multiple augmentations are precomputed and scored by the teacher using negative cross-entropy. We adopt a multi-round selection strategy where each image contributes at most one confident, unused augmentation per round, ensuring diversity in the initialized distilled set.
For each real image, multiple augmentations are precomputed and scored by the teacher model using negative cross-entropy. We adopt a multi-round selection strategy where each image contributes at most one augmentation per round, progressively selecting high-confidence variants to ensure diversity.
% To maintain batch consistency, zero-filled placeholders are inserted for tail classes to match the number of instances from those head-dominant ones.
To ensure consistent batch structure, we insert zero-filled placeholders for all classes having fewer instances than the largest one.

% Third, we introduce a confidence-aware, class-independent initialization strategy. For each real image, we generate multiple candidate variants and incrementally select high-confidence crops. To ensure sample-level diversity, each image is allowed to contribute at most once per round, and additional crops are selected only after all images in the same class have contributed. For convenience, we insert zero-filled placeholders to maintain batch consistency for all classes except the most head-dominant one.

Our main contributions are as follows:

\begin{itemize}
% \item We rethink long‑tailed dataset distillation via a uni‑level statistical alignment framework and directly tackle representation bias and skewed BN statistics.
\item We rethink long-tailed dataset distillation by 
moving from bi-level trajectory matching to a uni-level statistical alignment framework that better supports debiasing.
\item We implement unbiased recovery and soft relabeling through three key strategies: expert model debiasing; fair BN statistics recalibration; and confidence‑guided, multi‑round synthetic data initialization.
% \item We rethink long-tailed dataset distillation from the perspective of statistical alignment, aiming to improve accuracy by mitigating both model and statistic bias.
% \item Our uni-level framework improves representation alignment and soft label quality, yields more accurate and fair BN statistics, and constructs confidence-guided multi-round initialization tailored for long-tailed settings.
\item Extensive experiments on CIFAR-10-LT, CIFAR-100-LT, Tiny-ImageNet-LT, and ImageNet-LT demonstrate 
% that our method consistently outperforms 
our consistent superiority against state-of-the-art baselines.
% by large margins in under various long-tailed settings.
It improves accuracy by 15.6\% on CIFAR-100-LT and 11.8\% on Tiny-ImageNet-LT (IPC=10, IF=10).
% under the IPC=10, IF=10 setting.
\end{itemize}

\section{Related Work}
\subsection{Dataset Distillation}
% Dataset distillation aims to synthesize a distilled dataset that approximates the training effect of the full dataset. 
%Early dataset distillation methods are based on coreset selection, such as K-Center~\cite{kcenter} and GraphCut~\cite{graphcut}, which directly select a subset of real data. 
Early dataset distillation methods, such as K-Center~\cite{kcenter} and GraphCut~\cite{graphcut}, %rely on coreset selection by 
%directly choosing a subset of real data.
select a subset of real data directly, which limits the expressiveness of the resulting distilled dataset.
%However, these methods do not generate new data and are limited in expressiveness. 
Subsequent methods fall into three major categories. 
Gradient-matching-based approaches~\cite{dream,wang2025edf} align gradients between real and distilled data but scale poorly due to high memory usage. Trajectory-matching-based methods~\cite{cazenavette2022datasetmtt,mttnew} simulate training dynamics but are computationally expensive and memory-intensive. Distribution-matching-based methods~\cite{dm,optical} speed up convergence by matching features but still suffer from high memory costs and degrade on larger datasets like Tiny-ImageNet or ImageNet.
Recent efforts attempt to reduce memory overhead via generative-model-based approaches~\cite{otg,igd} or by adopting uni-level optimization~\cite{rded,edc}. %However, generative methods rely on pretraining on balancing large-scale datasets, while existing uni-level methods assume balanced data and lack debiasing mechanisms.
However, generative approaches typically rely on pretrained generators trained on balanced large-scale datasets, while existing uni-level methods operate under balanced assumptions and lack explicit debiasing strategies.
DAMED~\cite{damed} is the only prior work explicitly targeting long-tailed DD. However, it inherits representation bias from long-tailed expert training and retains the inefficiencies of trajectory-matching frameworks.
In contrast, our work is the first to systematically address long-tailed DD within a uni-level framework, with principled strategies for expert debiasing, image initialization, and unbiased alignment.

\subsection{Long-tailed Recognition}
% Long-tailed recognition refers to visual tasks performed under imbalanced data distributions, where a few head classes contain abundant samples while many tail classes have only limited examples~\cite{zhang2025systematic}. 
% To alleviate the resulting bias in representation learning, data augmentation strategies have been extensively explored~\cite{aug,aug2}. For example, Mixup~\cite{mixup} and its class-aware extension UniMix~\cite{unimix} encourage feature interpolation between samples to enrich tail-class supervision. CMO~\cite{cmo} further enhances this by generating context-aware mixed samples that better preserve semantic coherence for rare classes.
% Beyond augmentation, other efforts target network-level rebalancing through gradient adjustment~\cite{GR,GR2,GR3}, data synthesis via generative models or tailored instance generation~\cite{diffult,DS,ltgc2024}, and loss reweighting techniques~\cite{loss,loss2,loss3,du2023global} that amplify learning signals from underrepresented classes.
Long-tailed recognition refers to visual tasks performed under imbalanced data distributions~\cite{zhang2025systematic}. %, where a few head classes have abundant samples while many tail classes contain only limited examples~\cite{zhang2025systematic}.
To mitigate the resulting representation bias, data augmentation strategies have been widely studied~\cite{aug,aug2,li2025conmix}. For instance, Mixup~\cite{mixup} and its class-aware extension UniMix~\cite{unimix} promote feature interpolation to enrich supervision for tail classes, while CMO~\cite{cmo} generates context-aware mixed samples that better preserve semantic coherence for rare categories.
Beyond augmentation, other approaches mitigate long-tailed bias through network-level optimization~\cite{GR,GR2,GR3}, %(e.g., gradient reweighting~\cite{GR,GR2,GR3}), 
data synthesis using generative models or instance composition~\cite{diffult,DS,ltgc2024}, and loss rebalancing strategies~\cite{loss,loss2,loss3,du2023global} that amplify learning signals from underrepresented classes.
% Beyond augmentation, other approaches address long-tailed bias via network-level optimization, such as gradient reweighting~\cite{GR,GR2,GR3}, data synthesis using generative models or instance composition~\cite{diffult,DS,ltgc2024}, and loss rebalancing strategies~\cite{loss,loss2,loss3,du2023global} that amplify learning signals from underrepresented classes.
Given the limited attention to long-tailed dataset distillation, we draw conceptual insights from the broader long-tailed recognition literature to debias both the observer and teacher models, thereby enabling effective distillation under severe class imbalance.

\section{Methods}
\subsection{Problem Statement}
Long-tailed dataset distillation aims to construct a small, class-balanced synthetic dataset $\mathcal{S} = \{ (\bm{x}_s^i, y_s^i,\tilde{\mathbf{y}}_s^i) \}_{i=1}^{|\mathcal{S}|}$  from a long-tailed real dataset $\mathcal{D} = \{ (\bm{x}^i, y^i) \}_{i=1}^{|\mathcal{D}|}$, where $y^i \in \{0, \dots, C{-}1\}$ and $|\mathcal{S}| \ll |\mathcal{D}|$, such that models trained on $\mathcal{S}$ achieve strong generalization performance on a class-balanced real test set $\mathcal{T}$. Here, $y_s^i$ and $\tilde{\mathbf{y}}_s^i$ denote the hard and soft labels, respectively. Let $\mathcal{D}_c = \{ (\bm{x}^i, y^i) \in \mathcal{D} \mid y^i = c \}$ denote the subset of class $c$, where $|\mathcal{D}_0| > |\mathcal{D}_1| > \cdots > |\mathcal{D}_{C-1}|$ and $|\mathcal{D}_0| \gg |\mathcal{D}_{C-1}|$. The synthetic (distilled) dataset is class-balanced such that $\mathcal{S}_c = \{ (\bm{x}_s^i, y_s^i, \tilde{\mathbf{y}}_s^i) \in \mathcal{S} \mid y_s^i = c \}$ satisfies $|\mathcal{S}_0| = |\mathcal{S}_1| = \cdots = |\mathcal{S}_{C-1}|=\text{IPC}$. %A balanced test set $\mathcal{T} = \{ (\bm{x}_t^i, y_t^i) \}_{i=1}^{|\mathcal{T}|}$ is used to evaluate the generalization of student models trained on $\mathcal{S}$. 

\begin{figure}
    \centering
    \includegraphics[width=\linewidth]{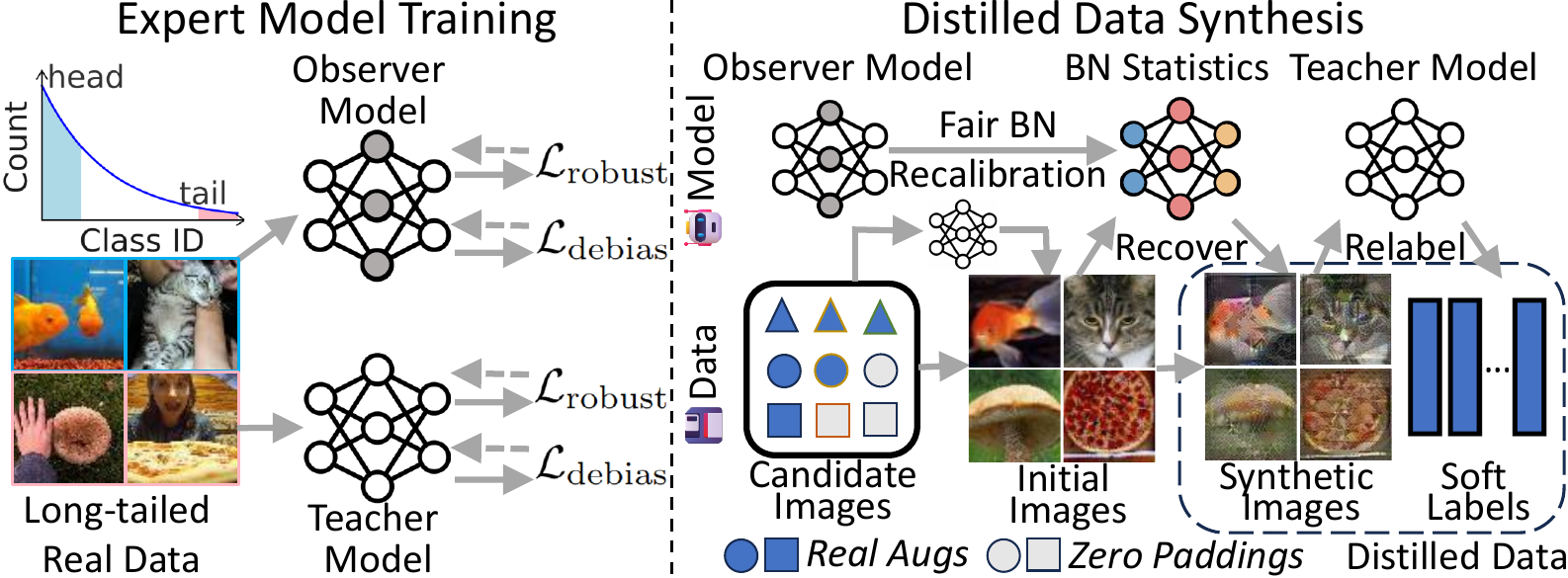}
    \caption{
    Illustration of our pipeline. We first train robust and debiased observer and teacher models on real data to enable reliable statistical alignment and semantic guidance. A candidate image pool tailored for long-tailed distributions is then used for teacher-guided initialization. Synthetic images are subsequently recovered via fair BN alignment and relabeled using teacher model to ensure semantic consistency.
    }
    \label{figure2}
\end{figure}

% Motivated by the limitations of trajectory-matching-based methods, we propose a uni-level statistical alignment framework for unbiased recovery and soft relabeling. This is achieved through expert model debiasing, BN statistics recalibration, and confidence-guided data initialization. The whole pipeline is shown in Fig.~\ref{figure2}.

To address the problem, we propose a uni-level statistical alignment framework for unbiased recovery and soft relabeling. Our approach departs from traditional trajectory-matching methods, whose fundamental limitations are detailed in the subsequent section. The success of our framework relies on three core strategies: expert model debiasing, fair BN statistics recalibration, and confidence-guided data initialization. The entire pipeline is illustrated in  Fig.~\ref{figure2}.

\subsection{Drawback of %DAMED and 
Trajectory-Matching-Based Methods}
% Trajectory-based dataset distillation methods typically begin by training an expert model $\theta_R$ on the real dataset $\mathcal{D}$, followed by alternating updates to a student model $\theta_S$ and a synthetic dataset $\mathcal{S}$ by minimizing the following objective:
Trajectory-based dataset distillation methods typically begin by training an expert model $\theta_R$ on the real dataset $\mathcal{D}$, followed by a bi-level optimization process, where the trajectory student model $\theta_S$ is trained on synthetic data, and the synthetic dataset $\mathcal{S}$ is updated by matching its induced trajectory to that of the expert $\theta_R$.
This objective is often formulated as minimizing the discrepancy between the training trajectories of the student and expert models:
% \begin{equation}
%  \mathcal{L}\left(\mathcal{S}\right) =   \mathbf{D}(\mathcal{S},\mathcal{D};\theta_S ,\theta_R)=\|F(\mathcal{S},\theta_S)-F(\mathcal{D}, \theta_R)\|_2.
% \end{equation}
\begin{equation}
 \mathcal{L}\left(\mathcal{S}\right) =\|F(\mathcal{S},\theta_S)-F(\mathcal{D}, \theta_R)\|_2,
\end{equation}
where $F$ denotes training trajectories.
However, when the expert model $\theta_R$ is trained on a long-tailed dataset,
% it inevitably encodes a class imbalance in
its internal representations are inevitably prone to class imbalance if no proper intervention is enforced.
Optimizing the student model to mimic such an expert causes the distilled dataset to inherit the bias, leading to over-emphasis on head-class semantics and
% limited coverage of 
under-representation of minority ones.

Although DAMED~\cite{damed} attempts to simulate imbalanced training dynamics within the student to reduce trajectory mismatch, it relies on an representation expert trained on imbalanced data without debiasing. As a result, the distilled dataset inherits the expert’s representational bias.
More broadly, the trajectory-based methods are difficult to balance between explicit debiasing and strict trajectory matching.
% explicitly debiasing inherently difficult.
Pre-distillation adjustments such as reweighting or logit correction %, or margin-aware objectives 
alter the expert’s optimization path, breaking the premise of trajectory matching.
Meanwhile, post-hoc debiasing is impractical, as these methods %are designed to 
only reproduce parameter evolution and lack fine-grained control over per-class representation quality.
In addition to these limitations, such methods incur substantial computational overhead due to the bi-level nature of the optimization, multi-step training trajectory simulation, and backpropagation through the unrolled computation graph.
\subsection{Uni-level Long-tailed Dataset Distillation}
% To overcome the limitations of trajectory-based approaches, we adopt a uni-level distillation framework that bypasses student trajectory simulation. This design not only reduces the computational burden but also enables explicit debiasing.
% As motivated in the introduction, we adopt a 
% uni-level distillation framework that 
Our uni-level distillation framework
avoids trajectory simulation and instead enables explicit debiasing through %two complementary 
%components: 
unbiased synthetic image recovery and unbiased soft relabeling. 
%We first describe how unbiased recovery is achieved. % in our framework.
%Specifically, 
To enable unbiased recovery of the synthetic images $\mathcal{S}$, we adopt a feature-statistics alignment objective to match their internal distribution to that of real data, without relying on repeated access to real samples during training.
%To this end,
Specifically, we leverage BN statistics as lightweight summaries of feature distributions.
We train an expert model $\theta_R$ (also referred to as the observer model) on real data and freeze it to serve as a stable source of BN statistics for both real and synthetic inputs.
The corresponding alignment loss is defined as:
\begin{equation}
\mathcal{L}\left(\mathcal{S}\right) =   \sum_{l=1}^{L}\mathbf{D}^{\mu}_l(\mathcal{S},\mathcal{D}; \theta_R)+\mathbf{D}^{\sigma}_l(\mathcal{S},\mathcal{D}; \theta_R), 
\label{object}
\end{equation}
where $l$ indexes the monitored BN layers, and $\mathbf{D}^{\mu}_l$, $\mathbf{D}^{\sigma}_l$ represent the $\ell_2$ distances between the mean ($\mu_l$) and variance ($\sigma_l$) of synthetic and real features in layer $l$:
\begin{equation}
\mathbf{D}^{\mu}_l(\mathcal{S},\mathcal{D}; \theta_R)=\left\| \mu_l(\mathcal{S}; \theta_R) - \mu_l(\mathcal{D}; \theta_R) \right\|_2 
\end{equation}
\begin{equation}
\mathbf{D}^{\sigma}_l(\mathcal{S},\mathcal{D}; \theta_R) = 
\left\| \sigma_l(\mathcal{S}; \theta_R) - \sigma_l(\mathcal{D}; \theta_R) \right\|_2 
\end{equation}
%As the above formulation shows, %the success of the statistical matching process relies on three key components for effective distillation in long-tailed scenarios: 
% The success of this recovery alignment %for recovery 
% under long-tailed settings hinges on three key factors:

The success of this alignment in long-tailed scenarios hinges on three factors:
(1) a robust and debiased observer model $\theta_R$
to serve as a reliable anchor for extracting statistics;
(2) fair and accurate
estimation of BN statistics ($\mu_l$, $\sigma_l$)
to handle both the intra-class and inter-class skew;
(3) a diverse and semantically 
meaningful initialization of $\mathcal{S}$
to effectively cover tail classes.
% While the observer model enables distribution-level consistency through BN statistics alignment, it does not directly enforce semantic discriminability. To address this, we incorporate an unbiased teacher model $\theta_T$ that provides soft labels for the synthetic images, complementing the statistical alignment with semantic supervision. 
%While the 
Although unbiased recovery enforces distribution-level consistency, it does not directly enforce semantic discriminability. 
To address this, we incorporate soft relabeling via an unbiased teacher model $\theta_T$, which produces balanced and informative supervision for $\mathcal{S}$. The resulting soft labels capture inter-class relationships and provide dense guidance that is %especially 
critical under class imbalance.

\paragraph{Debiasing the Observer and Teacher Models}
To ensure effective supervision in long-tailed settings, both the observer model $\theta_R$ and the teacher model $\theta_T$ must be trained to mitigate class imbalance and enhance robustness. The observer model guides distribution-level alignment via BN statistics, while the teacher model provides soft label~\cite{sink,sinkd} supervision for semantic guidance. If either model is biased, the distilled dataset would inherit skewed feature distributions or inaccurate soft labels. We address these challenges through two complementary loss functions. %We therefore optimize both models using complementary objectives tailored to their roles.
% to enhance representation robustness and mitigate class imbalance.
% To provide effective supervision in the long-tailed setting, both the observer model $\theta_R$ and the teacher model $\theta_T$ must be trained to reduce class imbalance and improve feature robustness. The observer aligns distributional statistics via BN, while the teacher model provides soft label supervision for semantic guidance. Bias in either model may propagate to the distilled dataset as skewed features or unreliable labels. We therefore apply complementary objectives to enhance their representation robustness and mitigate class imbalance.

First, to enhance the robustness of both $\theta_R$ and $\theta_T$, we use a symmetric mixture consistency loss that aligns representations across different mixed-label augmented images:
\begin{equation}
\mathcal{L}_{\mathrm{robust}} =
- \sum_{i=1}^{2} \cos\left(\mathbf{z}_i, \; \mathrm{sg}(\mathbf{p}_{\bar{i}})\right), 
\end{equation}
% where $\mathbf{z}_i$ and $\mathbf{p}_i$ are the predictor and projection-head outputs for the $i$-th augmented view, respectively, and $\bar{i}$ denotes the opposite view index. 
where $\bar{i} = 3 - i$ denotes the index of the opposite view. For the $i$-th augmented view, $\mathbf{z}_i$ is obtained by projecting the encoder output through a shared linear layer that maps features into a representation space. $\mathbf{p}_i$ is further transformed from $\mathbf{z}_i$ via an additional prediction layer.
%The transformation consists of two lightweight modules: one mapping encoder outputs into a comparison-friendly space, and the other introducing asymmetry to prevent representational collapse.
The stop-gradient operator $\mathrm{sg}(\cdot)$ ensures one-sided alignment for each term.

Second, to explicitly debias class imbalance during training, we adopt a dynamically-weighted rebalancing strategy that gradually emphasizes minority classes:
\begin{equation}
\mathcal{L}_{\mathrm{debias}}
= \alpha \sum_{k=0}^{C-1} {\frac{-(r_k)^{-q}y_k \log p_k}{\sum_{j=0}^{C-1} (r_j)^{-q}}}\, - \beta\sum_{k=0}^{C-1} y_k \log p_k
,
\end{equation}
where $y_k$ is the mixed target probability, $p_k$ is the predicted softmax output, $r_k$ is the sample frequency of class $k$, and $q$ controls the sharpness of reweighting. We set $\alpha = (\tfrac{t}{T})^2$ and $\beta = 1 - \alpha$, where $t$ and $T$ denote the current and total training iterations. This schedule gradually shifts focus toward minority classes while preserving early-stage stability.
% where $\alpha = (\tfrac{t}{T})^2$, $ \beta = 1-\alpha$, $r_k$ denotes the sample frequency of class $k$, $q$ controls the sharpness of reweighting, and $t$ and $T$ represent the current and total training iterations, respectively. This schedule prevents premature overfitting to minority classes and progressively corrects class imbalance. 

%as training proceeds.

Both $\theta_R$ and $\theta_T$ are trained with the combined loss:
\begin{equation}
  \mathcal{L} = \gamma_1 \mathcal{L}_{\mathrm{robust}} + \gamma_2\mathcal{L}_{\mathrm{debias}} , 
\end{equation}
where $\gamma_1$ and $\gamma_2$ are scalar weights for balancing.
%that balance the contributions from the statistical consistency loss and the semantic debiasing objective, ensuring both robustness and fairness in the learned representations.

%However, under the standard BN update rule with fixed momentum, the estimation of running mean and variance becomes inherently biased---even within the same class. Specifically, the exponential moving average updates assign higher weight to early-seen samples and diminishing weight to later ones. As a result, different images contribute unequally to the final statistics, which is particularly problematic in long-tailed datasets where some classes may only have a handful of samples. In such cases, each image should contribute equally; otherwise, the few available tail samples may be underrepresented even in their own class statistics.
\paragraph{Fair Recalibration of BN Statistics}
Accurate and fair BN statistics are critical in our framework, as they serve as alignment targets for image recovery.
However, under the standard exponential moving average update with fixed momentum, the running estimates become biased due to unequal sample contributions: recent batches dominate the statistics, while earlier ones are quickly forgotten. 
This effect is particularly problematic in long-tailed settings, where each tail-class sample carries high representational value and must contribute equally to the accumulated statistics. These limitations motivate our post-hoc recalibration strategy.
% This issue is further exacerbated in long-tailed settings, where each tail-class sample carries significant representational value and thus requires equal contribution in the accumulated statistics.
% Standard BN, however, fails to guarantee such fairness due to its fixed update rule, motivating the need for post-hoc recalibration.

To address this, we recalibrate BN statistics for the observer model $\theta_R$ after training, using a fair, sample-balanced and class-balanced accumulation strategy. Specifically, we freeze model parameters and perform a full forward pass over the real dataset, processing one batch at a time. For each monitored BN layer $l$, the class-$c$ mean is updated at batch index $t$ using a dynamically adjusted momentum $\alpha_t^c$: %At each batch index $t$, the mean for class $c$ is updated via a dynamically adjusted momentum for each monitored BN layer $l$: %(we omit the layer index $l$ here for clarity):
\begin{equation}
\mu_{l,t}^c = (1 - \alpha_t^c) \cdot \mu_{l,t-1}^c + \alpha_t^c \cdot \hat{\mu}_{l,t}^c, \quad
\alpha_t^c = \frac{B_t^c}{N_{t-1}^c + B_t^c },
\end{equation}
where $\mu_{l,t}^c$ denotes the cumulative mean of class-$c$ samples up to step $t$, $\hat{\mu}_{l,t}^c$ is their batch mean, $B_t^c$ is the number of class-$c$ samples in that batch, and $N_{t-1}^c$ is the cumulative count before step $t$.
Unlike standard BN with fixed momentum, our formulation introduces a dynamic momentum coefficient $\alpha_t^c$, ensuring that each sample contributes equally to the final statistics regardless of its temporal order.

% where $\mu_t^c$ denotes the cumulative mean of class-$c$ samples up to step $t$, $\hat{\mu}_t^c$ is the mean and $B_t^c$ the count of class-$c$ samples in batch $t$, and $N_{t-1}^c$ is the cumulative count of class-$c$ samples before batch $t$. 

After the full pass, we aggregate global BN statistics by uniformly averaging across all class-specific means to ensure class-balanced statistics. Let $T$ denote the total number of batches in the forward pass. The global mean is given by:
\begin{equation}
\mu_l(\mathcal{D}; \theta_R) = \frac{1}{C} \sum_{c=0}^{C-1} \mu_{l,T}^c(\mathcal{D}; \theta_R),
\end{equation}
and similarly for variance $\sigma$. 
This two-stage strategy mitigates both intra-class bias and inter-class bias, thereby serving as reliable supervision signals for statistical alignment.

\paragraph{Confidence-guided Multi-round Initialization}
% Effective initialization of synthetic images facilitates stable optimization 
% and ensures comprehensive coverage,
% %and reliable alignment of BN statistics,
% particularly under long-tailed settings due to data imbalance. 
Initialization primarily governs the diversity of final synthetic images, while also playing a supportive role in promoting stable optimization under long-tailed distributions.
Traditional initialization strategies typically rely on either sampling real images or random noise. However, random initialization often leads to poor convergence and degraded downstream performance.
In highly imbalanced settings, directly sampling real images becomes infeasible, as tail classes often contain too few samples to provide adequate initialization.

To overcome this limitation, we propose a confidence-guided, multi-round initialization strategy tailored for long-tailed distributions. Specifically, we generate multiple augmentations (e.g., crops) per real image and score them via the teacher model $\theta_T$ using negative cross-entropy loss. %For each image, only the most confident augmentation is retained.
% If the number of real images exceeds the number of required synthetic samples, we directly select the top-ranked augmentations---each image contributes at most once.
% For each image, only its most confident unused augmentation is retained in the current round.
% When the number of confident candidates exceeds the per-class IPC, we select the top-ranked samples—--each from a distinct real image.
% Otherwise, when real images are insufficient, we perform multi-round selection: in each round, every image contributes its highest-confidence unused augmentation. This process continues until the target number of initializations is reached.
% Otherwise, we perform multi-round selection: in each round, every image contributes its most confident unused augmentation. If a class has enough candidates, we select the most confident ones to fill the remaining slots; otherwise, all are retained. This process repeats until all classes reach their target IPC.
These augmentations are stored in a class-wise candidate pool.
In each round, every real image contributes its most confident unused augmentation to a temporary selection pool. If the total number of candidates exceeds the remaining slots for that class, we select the top-scoring augmentations; otherwise, we retain all. This process repeats until each class reaches its target IPC.
This strategy ensures high-confidence selection while maintaining sample-level diversity across varying class sizes.
To maintain structural consistency across classes, we insert zero-initialized placeholders for classes with fewer real samples than the largest class. These placeholders are excluded from the augmentation and selection process, ensuring all synthetic samples are semantically meaningful. %See Appendix for the full algorithm.

\begin{table*}[t]
\centering
\small
\begin{tabular}{l|ccc|ccc|ccc}
\hline
Dataset &\multicolumn{9}{c}{CIFAR-10-LT} \\
\hline
Imbalance Factor & \multicolumn{3}{c|}{10} & \multicolumn{3}{c|}{50} & \multicolumn{3}{c}{100} \\
Images per Class & 10 & 20 & 50 & 10 & 20 & 50 & 10 & 20 & 50 \\
\hline
Random & 32.5$\pm$2.2 & 39.6$\pm$0.9 & 51.9$\pm$1.5 & 33.2$\pm$0.4 & 42.0$\pm$1.3 & 51.6$\pm$1.3 & 34.4$\pm$2.0 & 41.4$\pm$0.7 & 52.6$\pm$0.5 \\
K-Center (ICLR'18) & 21.9$\pm$0.8 & 24.2$\pm$0.8 & 31.7$\pm$0.9 & 17.8$\pm$0.2 & 20.8$\pm$0.5 & 26.1$\pm$0.2 & 16.2$\pm$0.5 & 19.0$\pm$1.0 & 24.2$\pm$1.2 \\
Graph-Cut (ALT'21) & 28.7$\pm$0.9 & 34.2$\pm$1.0 & 40.6$\pm$1.0 & 24.2$\pm$0.7 & 28.6$\pm$0.8 & 33.9$\pm$0.4 & 22.9$\pm$0.9 & 26.0$\pm$0.5 & 33.3$\pm$1.0 \\
\hline
DC (ICLR'21) & 37.9$\pm$0.9 & 38.5$\pm$0.9 & 37.4$\pm$1.4 & 37.3$\pm$0.9 & 38.8$\pm$1.0 & 35.8$\pm$1.2 & 36.7$\pm$0.8 & 38.1$\pm$1.0 & 35.3$\pm$1.4 \\
MTT (CVPR'22) & 58.0$\pm$0.8 & 59.5$\pm$0.4 & 62.0$\pm$0.9 & 45.8$\pm$1.4 & 49.9$\pm$0.8 & 53.6$\pm$0.5 & 37.7$\pm$0.6 & 41.6$\pm$1.1 & 47.8$\pm$1.1 \\
DREAM (ICCV'23) & 34.6$\pm$0.6 & 42.2$\pm$1.5 & 50.5$\pm$0.7 & 30.8$\pm$0.6 & 38.4$\pm$0.3 & 45.5$\pm$0.9 & 30.8$\pm$1.7 & 34.9$\pm$0.8 & 42.2$\pm$0.8 \\
IDM (CVPR'23) & 54.8$\pm$0.4 & 57.1$\pm$0.3 & 60.1$\pm$0.3 & 51.9$\pm$0.7 & 53.3$\pm$0.6 & 56.1$\pm$0.4 & 49.8$\pm$0.6 & 50.9$\pm$0.5 & 53.1$\pm$0.4 \\
Minimax (CVPR'24) & 29.2$\pm$0.5 & 28.5$\pm$0.6 & 39.9$\pm$0.1 & 18.4$\pm$0.3 & 22.5$\pm$0.2 & 25.2$\pm$0.2 & 19.9$\pm$0.4 & 23.3$\pm$0.2 & 28.0$\pm$0.6 \\
DATM (ICLR'24) & 57.2$\pm$0.4 & 60.4$\pm$0.2 & 66.7$\pm$0.6 & 41.6$\pm$0.2 & 43.4$\pm$0.3 & 50.3$\pm$0.2 & 37.3$\pm$0.2 & 38.9$\pm$0.1 & 44.3$\pm$0.1 \\
RDED* (CVPR'24) & 46.2$\pm$0.2 & 55.5$\pm$0.3 & 62.0$\pm$0.3 & 43.3$\pm$0.3 & 46.1$\pm$0.3 & 53.2$\pm$0.3 & 41.2$\pm$0.3 & 45.3$\pm$0.2 & 49.9$\pm$0.3 \\
EDC* (NeurIPS'24) & 56.4$\pm$0.2 & 62.7$\pm$0.3 & 68.5$\pm$0.1 & 55.2$\pm$0.2 & 60.2$\pm$0.2 & 64.1$\pm$0.3 & 53.2$\pm$0.5 & 57.4$\pm$0.3 & 60.5$\pm$0.4 \\
DAMED (CVPR'25) & 58.1$\pm$0.3 & 63.0$\pm$1.0 & 70.5$\pm$0.4 & 54.2$\pm$1.0 & 59.4$\pm$0.7 & 65.8$\pm$0.2 & 53.4$\pm$0.1 & 58.2$\pm$0.6 & 64.0$\pm$0.9 \\
Ours &\bf63.6$\pm$0.5&\bf68.3$\pm$0.2&\bf74.1$\pm$0.2&\bf62.9$\pm$0.2&\bf67.3$\pm$0.3&\bf70.6$\pm$0.1&\bf62.7$\pm$0.1&\bf66.4$\pm$0.2&\bf68.8$\pm$0.4\\
\hline
\end{tabular}
\caption{Quantitative comparisons on CIFAR-10-LT. Rows marked with * indicate experiments conducted using open-source implementations with adaptations. Other results are taken from DAMED~\cite{damed}. }
\label{tab:cifar10}
\end{table*}

\begin{table*}[htbp]
\centering
\footnotesize
\begin{tabular}{l|ccc|ccc|ccc}
\hline
Dataset         & \multicolumn{9}{c}{CIFAR-100-LT} \\
\hline
Imbalance Factor & \multicolumn{3}{c|}{10}       & \multicolumn{3}{c|}{20}        & \multicolumn{3}{c}{50}        \\
Images per Class              & 10           & 20           & 50           & 10           & 20           & 50           & 10           & 20           & 50           \\
\hline
Random           & 14.2$\pm$0.6  & 21.7$\pm$0.6  & 32.1$\pm$0.6  & 15.0$\pm$0.3   & 21.6$\pm$0.5   & 30.5$\pm$0.5   & 13.4$\pm$0.5   & 20.6$\pm$0.6   & 26.9$\pm$0.5   \\
K-Center (ICLR'18)  & 10.7$\pm$0.9  & 15.9$\pm$1.0  & 24.8$\pm$0.2  & 10.0$\pm$0.5   & 15.1$\pm$0.6   & 23.8$\pm$0.3   &  8.7$\pm$0.6   & 12.4$\pm$0.6   & 19.8$\pm$0.5   \\
Graph-Cut (ALT'21)       & 16.9$\pm$0.3  & 22.2$\pm$0.4  & 29.9$\pm$0.4  & 16.0$\pm$0.5   & 20.7$\pm$0.5   & 28.7$\pm$0.3   & 13.1$\pm$0.5   & 17.2$\pm$0.6   & 24.8$\pm$0.5   \\
\hline
DC (ICLR'21)          & 24.0$\pm$0.3  & 27.4$\pm$0.3  & 27.4$\pm$0.3  & 23.2$\pm$0.3   & 26.2$\pm$0.3   & 27.4$\pm$0.3   & 19.8$\pm$0.4   & 22.7$\pm$0.4   & 25.9$\pm$0.3   \\
MTT (CVPR'22)             & 14.3$\pm$0.1  & 16.7$\pm$0.2  & 13.8$\pm$0.2  & 12.6$\pm$0.3   & 15.0$\pm$0.2   & 10.6$\pm$0.5   &  8.2$\pm$0.5   & 11.2$\pm$0.5   &  6.3$\pm$0.5   \\
DREAM (ICCV'23)           & 10.1$\pm$0.4  & 12.0$\pm$1.0  & 13.1$\pm$0.4  &  9.4$\pm$0.4   & 10.3$\pm$0.6   & 12.3$\pm$0.3   &  6.8$\pm$0.5   &  7.6$\pm$0.5   &  9.8$\pm$0.5   \\
DATM (ICLR'24)      & 28.2$\pm$0.4  & 34.1$\pm$0.2  & 31.6$\pm$0.1  & 25.3$\pm$0.3   & 27.2$\pm$0.1   & 27.1$\pm$0.1   & 22.2$\pm$0.4   & 19.8$\pm$0.4   & 22.9$\pm$0.4   \\
RDED* (CVPR'24) &30.5$\pm$0.4&36.0$\pm$0.2&37.6$\pm$0.1&32.1$\pm$0.3&33.9$\pm$0.2&-&28.0$\pm$0.2&-&- \\
EDC* (NeurIPS'24) &31.5$\pm$0.3&36.6$\pm$0.1&38.5$\pm$0.2&32.7$\pm$0.3&34.2$\pm$0.2&-&30.2$\pm$0.2&-&- \\
DAMED (CVPR'25)      & 31.5$\pm$0.2  & 37.5$\pm$0.4  & 40.0$\pm$0.1  & 31.4$\pm$0.5   & 35.1$\pm$0.4   & 37.0$\pm$0.7   & 29.8$\pm$0.3   & 31.9$\pm$0.5   & 33.2$\pm$0.5   \\
Ours             & \bf47.1$\pm$0.1  & \bf48.8$\pm$0.1  & \bf49.9$\pm$0.3  & \bf45.5$\pm$0.4   & \bf46.9$\pm$0.2   & \bf48.1$\pm$0.2   & \bf42.1$\pm$0.1   & \bf43.4$\pm$0.2   & \bf44.2$\pm$0.1   \\
\hline
\end{tabular}
\caption{Quantitative comparisons on CIFAR-100-LT. Rows marked with * indicate experiments conducted using open-source code with adaptations. Other results are taken from DAMED~\cite{damed}. 
‘–’ means distillation could not be executed. }
\label{tab:cifar100}
\end{table*}

\begin{table}[htb]
\centering
\footnotesize
\begin{tabular}{l|cc|cc}
\hline
Dataset & \multicolumn{4}{c}{Tiny-ImageNet-LT} \\
\hline
IF & \multicolumn{2}{c|}{10} & \multicolumn{2}{c}{20} 
\\
%\hline
IPC & 10 & 20 & 10 & 20 \\
\hline
Random & 7.4$\pm$0.2 & 13.5$\pm$0.4 & 7.6$\pm$0.1 & 13.2$\pm$0.4 \\
K-Center & 10.4$\pm$1.5 & 11.3$\pm$1.2 & 12.6$\pm$1.4 & 9.9$\pm$2.2 \\
Graph-Cut & 9.8$\pm$0.7 & 5.6$\pm$0.6 & 3.4$\pm$0.8 & 5.0$\pm$1.0 \\
\hline
MTT & 11.1$\pm$0.2 & 18.1$\pm$0.2 & 7.7$\pm$0.1 & 14.7$\pm$0.2 \\
DREAM & 5.4$\pm$0.3 & 6.8$\pm$0.1 & 4.8$\pm$0.1 & 6.0$\pm$0.2 \\
DATM & 21.3$\pm$0.1 & 14.5$\pm$0.1 & 14.0$\pm$0.6 & 19.0$\pm$0.3 \\
RDED* &22.7$\pm$0.3&23.3$\pm$0.2&21.0$\pm$0.3&21.8$\pm$0.3 \\
EDC* &23.9$\pm$0.6&25.2$\pm$0.2&22.1$\pm$0.5&23.7$\pm$0.4 \\
DAMED & 26.0$\pm$0.3 & 27.9$\pm$0.2 & 23.6$\pm$0.3 & 25.5$\pm$0.3 \\
Ours&\bf37.8$\pm$0.4&\bf38.9$\pm$0.2&\bf36.1$\pm$0.1&\bf37.0$\pm$0.1\\
\hline
\end{tabular}
\caption{Quantitative comparisons on Tiny-ImageNet-LT. Rows marked with * indicate experiments conducted using open-source implementations with adaptations. Other baseline results are taken from DAMED~\cite{damed}.
}
\label{tab:tinyimagenet}
\end{table}

\paragraph{Recovery, Relabeling and Evaluation}
Given an initialized distilled set $\mathcal{S}_{\text{init}}$, an observer model $\theta_R$, a teacher model $\theta_T$, and precomputed BN statistics from the real dataset, 
we optimize the distilled set $\mathcal{S}$ (initialized from $\mathcal{S}_{\text{init}}$) through statistical recovery using $\theta_R$ and real-data BN statistics ($\mu_l(\mathcal{D};\theta_R)$, $\sigma_l(\mathcal{D},\theta_R)$), followed by soft relabeling via $\theta_T$.
% we optimize the synthetic dataset $\mathcal{S}$ (initialized from $\mathcal{S}_{\text{init}}$) through a two-stream supervision process: statistical alignment recovery and semantic relabeling.

For statistical recovery, we pass all images in $\mathcal{S}$ through the observer model $\theta_R$ to compute per-layer BN statistics $\mu_l(\mathcal{S}; \theta_R)$ and $\sigma_l(\mathcal{S}; \theta_R)$, and align them with the precomputed real-data statistics $\mu_l(\mathcal{D}; \theta_R)$ and $\sigma_l(\mathcal{D}; \theta_R)$ as defined in Eq.~\eqref{object}. 
In practice, we also perform class-wise alignment by computing statistics separately for each class in $\mathcal{S}$ and matching them to their corresponding real-data counterparts.

% Once recovery is complete, we relabel each synthetic image by passing it through the trained, unbiased $\theta_T$. For each synthetic image $\bm{x}_s^i$, we obtain a soft label $\tilde{\mathbf{y}}_s^i = \theta_T(\bm{x}_s^i)$.  Meanwhile, we retain the original one-hot label $y_i$ for matching. %from initialization to ensure class identity.
Once recovery is complete, each synthetic image is relabeled by passing it through $\theta_T$. Specifically, for each synthetic image $\bm{x}_s^i$, we obtain a soft label $\tilde{\mathbf{y}}_s^i = \theta_T(\bm{x}_s^i)$, while retaining the original one-hot label $y^i_s$ for supervision.

To evaluate the quality of %the distilled dataset 
$\mathcal{S}$, we follow~\cite{edc} and train a student model $s$ from scratch on $\mathcal{S}$ using a dual-objective loss that combines one-hot labels and soft targets:
\begin{equation}
\mathcal{L}_{\text{match}} = \kappa_1 \cdot \mathcal{L}_{\text{CE}}( s(\bm{x}_s^i),y_s^i) + \kappa_2 \cdot \|\tilde{\mathbf{y}}_s^i - s(\bm{x}_s^i)\|_2^2,
\end{equation}
where $\mathcal{L}_{\text{CE}}(\cdot,\cdot)$ denotes the cross-entropy loss, $\|\cdot\|_2$ represents $l_2$ distance, $\kappa_1$ and $\kappa_2$ are weights for balancing.

\section{Experiments}
\subsection{Experimental Settings}
\paragraph{Datasets} To comprehensively evaluate the effectiveness of our method under varying degrees of class imbalance, we conduct experiments on four long-tailed benchmarks across different scales: CIFAR-10-LT, CIFAR-100-LT~\cite{cui2019classcifarlt}, Tiny-ImageNet-LT~\cite{tiny-imagenet}, and ImageNet-LT~\cite{liu2019largeimagenetlt}. These datasets are constructed by applying an exponential decay sampling strategy to the balanced  datasets, following the protocols of prior works~\cite{damed, cui2019classcifarlt}. Specifically, the number of samples for class $c$ is determined by $|\mathcal{D}_c| = |\mathcal{D}_0| \phi^c$, where $\phi^c = \beta^{-(c/(C-1))}$, with $\beta$ representing the imbalance factor (IF). A larger value of $\beta$ indicates more severe class imbalance. We evaluate under multiple imbalance ratios and varying images-per-class (IPC) budgets to reflect both mild and extreme long-tailed settings.

\paragraph{Network Architectures}
Following the settings established in 
DAMED~\cite{damed}, we adopt a depth-3 ConvNet as the student model for CIFAR-10-LT and CIFAR-100-LT, and a depth-4 ConvNet for Tiny-ImageNet-LT and ImageNet-LT. Given the superior representational capacity of ResNet architectures for large-scale datasets, we additionally evaluate ResNet-50 on ImageNet-LT under highly imbalanced scenarios. 
% Given ResNet-50's superior representational capacity to handle large-scale data, we additionally evaluate it on ImageNet-LT under severe class imbalance.
During evaluation, all student models are trained for 1000 epochs on the distilled dataset. All experiments were repeated five times for fairness, and conducted primarily on a single NVIDIA RTX 3090 GPU. %All experiments were conducted with 5 evaluations for fairness, primarily using a single NVIDIA 3090 GPU.

\paragraph{Baselines} We compare our method with a diverse set of representative baselines, including coreset selection methods such as Random, K-Center Greedy~\cite{kcenter}, and Graph Cut~\cite{graphcut}; gradient-matching-based methods including DC~\cite{dc} and DREAM~\cite{dream}; distribution-matching-based methods such as CAFE~\cite{cafe} and IDM~\cite{zhao2023improvedidm}; trajectory-matching-based methods including MTT~\cite{cazenavette2022datasetmtt}, DATM~\cite{datm}, TESLA~\cite{tesla}, and DAMED~\cite{damed}; uni-level optimization methods including SRe$^2$L~\cite{sre2l}, RDED~\cite{rded} and EDC~\cite{edc}; and generative-model-based methods such as Minimax~\cite{gu2024efficientminimax}. %Details of our hyperparameter settings, tuning protocol, and sensitivity analysis are provided in the appendix for reproducibility.
%To ensure compatibility under long-tailed settings, we extend RDED and EDC by incorporating structrual blanks for initialization.

\subsection{Results and Discussions}

\begin{table}[tb]
\centering
\footnotesize
\begin{tabular}{l|cc|cc}
\hline
Dataset & \multicolumn{4}{c}{ImageNet-LT} \\
\hline
IF & \multicolumn{2}{c|}{5} & \multicolumn{2}{c}{10} \\
%\hline
IPC & 10 & 20 & 10 & 20 \\
\hline
Random & 3.9$\pm$0.1 & 7.0$\pm$0.1 & 3.9$\pm$0.1 & 6.8$\pm$0.1 \\
G-VBSM & - & 1.0$\pm$0.1 & - & 1.0$\pm$0.1 \\
TESLA & 3.0$\pm$0.1 & - & 2.7$\pm$0.1 & -  \\
DATM & 7.4$\pm$0.1 & 8.1$\pm$0.2 & 7.9$\pm$0.1 & 8.2$\pm$0.1  \\
SRe$^2$L & 6.7$\pm$0.1 & 10.1$\pm$0.1 & 7.7$\pm$0.3 & 10.9$\pm$1.0 \\
DAMED & 20.8$\pm$0.2 & 21.0$\pm$0.1 & 20.3$\pm$0.1 & 20.7$\pm$0.1 \\
Ours &\bf24.7$\pm$0.4&\bf25.0$\pm$0.3&\bf23.5$\pm$0.2&\bf24.1$\pm$0.4 \\
\hline
% Full Dataset & \multicolumn{2}{c|}{27.6$\pm$0.3} & \multicolumn{2}{c}{26.6$\pm$0.2} \\
% \hline
\end{tabular}
\caption{Quantitative comparisons on ImageNet-LT. Baseline results are taken from DAMED~\cite{damed}.}
\label{tab:imagenet}
\end{table}

\begin{table}[tb]
    \centering
    \small
    \begin{tabular}{l|cc|cc}
    \hline
    Dataset & \multicolumn{2}{c|}{Tiny-ImageNet-LT} &\multicolumn{2}{c}{ImageNet-LT (ResNet)} \\
\hline
IF & \multicolumn{2}{c|}{100} & \multicolumn{2}{c}{256} \\
%\hline
      IPC   &  10& 20 &  10& 20\\
      \hline
      MTT &5.6$\pm$0.2&7.7$\pm$0.3&-&- \\
      DATM   &10.1$\pm$0.3  &11.9$\pm$0.1 &6.2$\pm$0.3 &6.7$\pm$0.2\\
      DAMED &17.1$\pm$0.8&18.0$\pm$0.5&17.2$\pm$0.2 &17.9$\pm$0.2 \\
      Ours &\bf28.5$\pm$0.2&\bf29.8$\pm$0.1&\bf48.2$\pm$0.7&\bf 48.9$\pm$0.2 \\
      \hline
    \end{tabular}
    \caption{Quantitative comparisons under severe class imbalance, with ResNet-50 used for ImageNet-LT evaluation.}
    \label{imbalanced}
\end{table}

\begin{figure}[tb]
    \centering \includegraphics[width=0.48\linewidth]{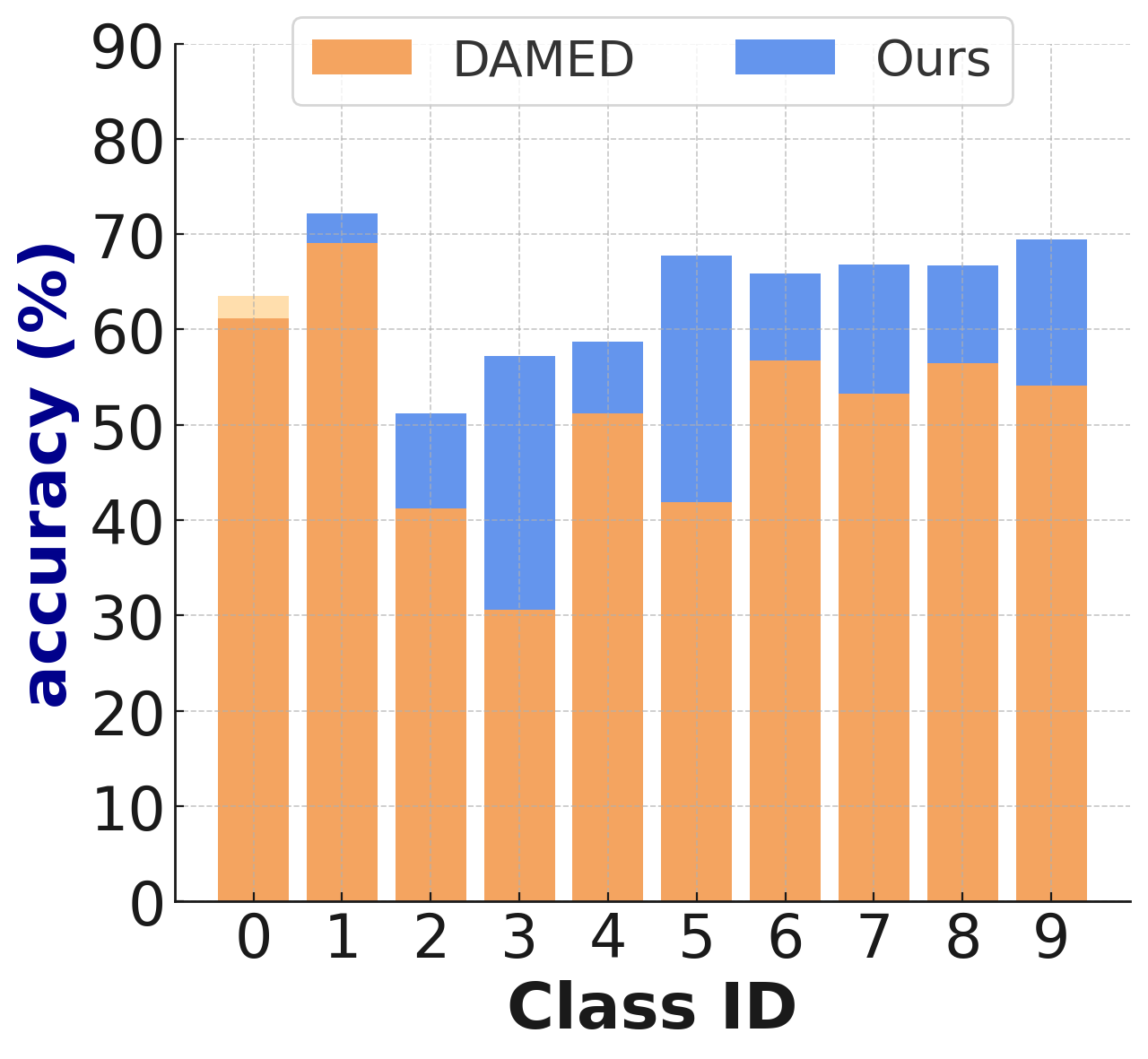}
\includegraphics[width=0.48\linewidth]{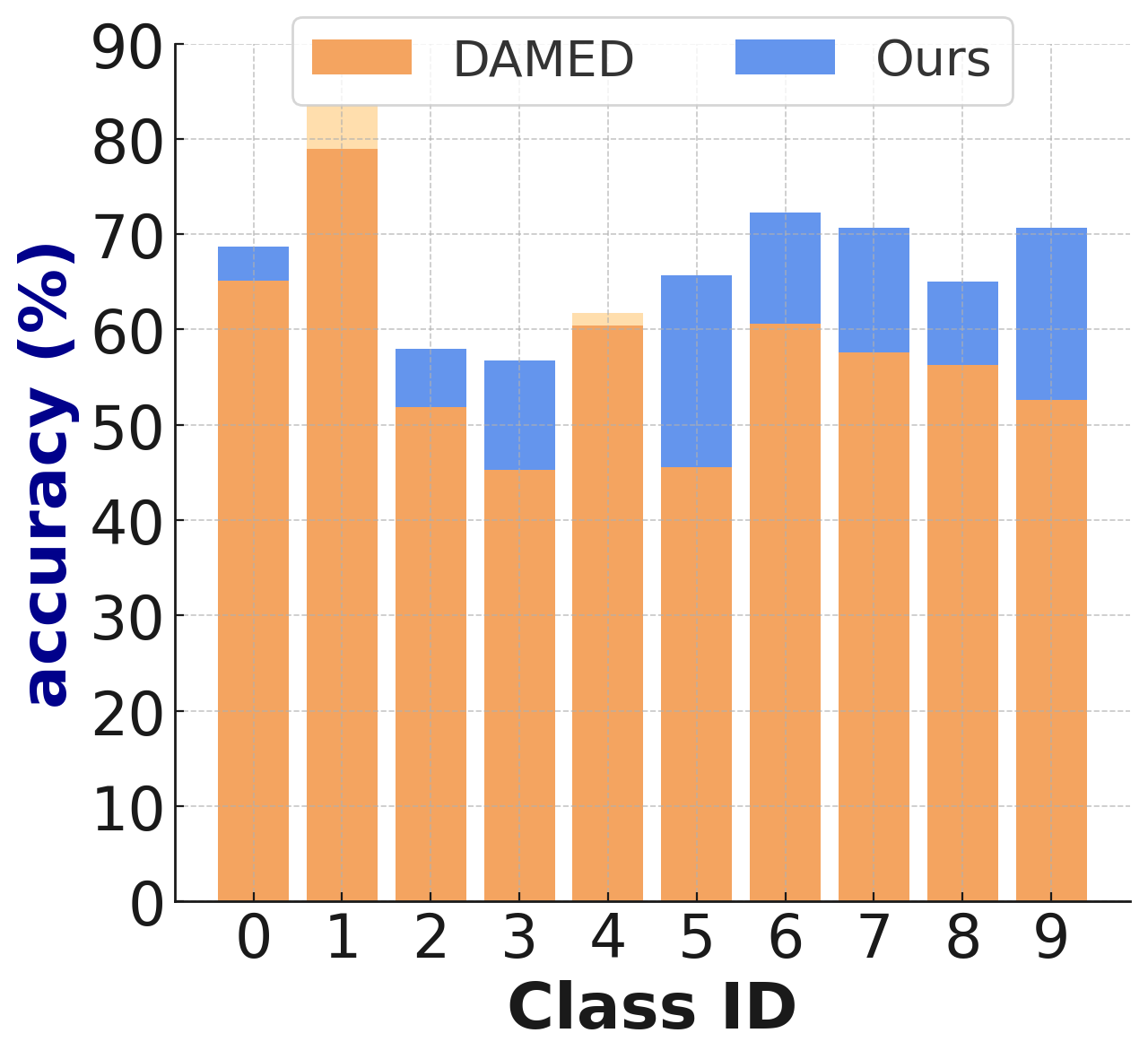}
\vspace{-0.5em}
    \caption{Class-wise accuracy comparison on CIFAR-10-LT (IF = 100). The left figure corresponds to IPC = 10, and the right figure to IPC = 20. For both settings, we compare our method with DAMED~\cite{damed}.}
    \label{classwise}
\end{figure}

\begin{table}[tb]
    \centering
    \small
    \begin{tabular}{l|c|c|c}
    \hline
    Dataset & CIFAR-10-LT &CIFAR-100-LT & Tiny-IN-LT \\
\hline
IF & 100 & 50 & 100 \\
      %IPC   &  1 &1&1\\
      \hline
      CAFE &15.2$\pm$0.8&11.3$\pm$0.9&- \\
      DREAM &15.3$\pm$0.7&11.7$\pm$0.7&1.9$\pm$0.2 \\
      DATM   &20.4$\pm$0.5  &12.8$\pm$0.4 &5.5$\pm$0.2 \\
      G-VBSM &4.1$\pm$0.2&1.8$\pm$0.2&2.3$\pm$0.1 \\
      RDED &21.8$\pm$0.4&15.6$\pm$0.2&9.1$\pm$0.3 \\
      DAMED &24.1$\pm$0.5&7.8$\pm$0.3&6.0$\pm$0.3 \\
      Ours &\bf44.8$\pm$0.3
      %40.2$\pm$0.1
      &\bf31.8$\pm$0.1&\bf20.1$\pm$0.2  \\
      \hline
    \end{tabular}
    \caption{Performance comparison under extreme settings with only 1 image per class (IPC = 1).}
    \label{ipc}
\end{table}

\begin{table}[t]
    \centering
    \small
    \begin{tabular}{c|cccc}
    \hline
      Method & ConvNet-3 & VGG-11 & ResNet-18 & AlexNet \\
      \hline
      MTT &37.7&27.4&29.6&26.3 \\
      DATM &37.3&34.4&35.1&36.8 \\
      RDED &41.2&41.6&37.5&28.2 \\
      DAMED &53.4&29.7&43.6&37.9 \\
      Ours &\bf62.7&\bf64.6&\bf58.7&\bf56.5 \\
      \hline
    \end{tabular}
    \caption{Cross-architecture evaluation on CIFAR-10-LT under an imbalance factor of 100, using 10 images per class.
    %Cross-architecture evaluation on CIFAR-10-LT under an imbalance factor of 100 and 10 images per class.
    }
    \label{cross}
\end{table}

\begin{table}[tb]
    \centering
    \small
    \begin{tabular}{c|ccc}
    \hline
      Method   & IPC=10 & IPC=20 & IPC=50 \\
         \hline
     w/o Model Debiasing   & 31.7&32.3&32.8  \\
     w/o Statistics Recalibration &40.9&41.8&42.1  \\
     w/o Adapted Initialization&40.8&-&- \\
     Full &\bf42.1&\bf43.4&\bf44.2 \\
     \hline
    \end{tabular}
    \caption{Ablation study on CIFAR-100-LT (IF=50).}
    \label{ablation}
\end{table}

% Such settings severely challenge the distilled set's capacity to represent the underlying data distribution: with minimal capacity per class, it becomes difficult to preserve intra-class diversity, ensure stable optimization, or provide meaningful supervision for the student model. 

\begin{table}[t]
    \centering
    \small
    \begin{tabular}{c|cc|cc}
    \hline
      Dataset & \multicolumn{2}{c|}{CIFAR-10-LT (IF=100)} & \multicolumn{2}{c}{CIFAR-100-LT (IF=50)} \\
    %\hline
    Stage & EMT & DDS & EMT & DDS \\
    \hline
    DAMED &31388s&30141s&26269s&29328s \\
    Ours  &\bf2395s&\bf118s&\bf2183s&\bf273s \\
    \hline
    \end{tabular}
    \caption{Runtime comparison on expert model training (EMT) and distilled data synthesis (DDS) between DAMED~\cite{damed} and our approach (IPC=1).}
    \label{runtime}
\end{table}

\begin{table}[tb]
    \centering
    \small
    \begin{tabular}{c|c@{\hskip 10pt}c@{\hskip 10pt}c|cc}
      \hline
      \multirow{2}{*}{Dataset}   & \multicolumn{3}{c|}{CIFAR-100-LT} & \multicolumn{2}{c}{Tiny-ImageNet-LT} \\
      &IPC=10&IPC=20&IPC=50&IPC=1&IPC=10 \\
         \hline
    DAMED &10.2&12.1&15.8&10.2&$>$24.0 \\
    Ours &\bf3.1&\bf3.1&\bf3.1&\bf6.1&\bf6.1 \\
    \hline
    \end{tabular}
    \caption{Peak GPU memory (GB)
comparison between DAMED~\cite{damed} and our approach.}
    \label{memory}
\end{table}

\paragraph{Main Results} We conduct comprehensive evaluations covering a broad spectrum of IF and IPC configurations, covering datasets of varying complexity. As shown in Tables.~\ref{tab:cifar10}, \ref{tab:cifar100}, \ref{tab:tinyimagenet}, and \ref{tab:imagenet}, our method consistently outperforms strong baselines under all evaluated settings.
While DAMED~\cite{damed} yields student performance closely matching those of its biased expert—--effectively saturating its performance ceiling—--our method explicitly mitigates expert bias, enabling the distilled data to supervise more accurate and generalizable student models, thereby raising the achievable upper bound.
% Although DAMED can achieve similar performance with their expert models in many settings, our method significantly elevate the performace ceiling by the explicit mitigation of long-tailed biases and the enhanced representational accuracy it brings. 
By aligning model representations in a class-balanced manner and recalibrating balanced BN statistics, our approach avoids the typical overfitting to head classes and promotes equitable learning across classes and training samples.
%across the class spectrum. 
Our debiasing mechanisms allow the distilled dataset to preserve both structural fidelity and semantic completeness, making our method broadly effective across datasets of different scales. %and resolutions.

\paragraph{Results under Highly Imbalanced Settings}
% Table~\ref{imbalanced} summarizes the results under highly imbalanced scenarios with large imbalance factors. These settings pose substantial challenges for dataset distillation, especially when the number of available real images is smaller than the target IPC for certain classes. Under such constraints, prior methods like RDED~\cite{rded} and EDC~\cite{edc} are not applicable to due to insufficient real data in tail classes. Other baselines also cannot achieve good results caused by biased representation learning or optimization instability.
Table~\ref{imbalanced} summarizes the results under highly imbalanced scenarios. %with large IF. 
These settings pose significant challenges for dataset distillation, particularly when the number of available real images falls below the target IPC for certain classes. 
% Under such constraints, prior methods such as RDED~\cite{rded} and EDC~\cite{edc} become inapplicable due to the scarcity of tail-class samples for initialization.
% Under such constraints, prior methods such as RDED~\cite{rded} and EDC~\cite{edc} become inapplicable, as they rely on sufficient tail-class samples.
Under such constraints, some prior methods become inapplicable; for example, EDC’s initialization and RDED’s sampling mechanism fail due to insufficient tail‑class examples.
Other baselines also struggle to achieve competitive performance, often due to biased representation learning or optimization instability.
In contrast, our method consistently achieves stronger performance across all tested configurations. Notably, under IF = 256 with ResNet-50 as the evaluation model, our distilled set for ImageNet-LT~%\cite{liu2019largeimagenetlt} 
not only outperforms those generated by all competing methods under the same imbalanced setting, but also surpasses several methods whose distilled sets are obtained using the full, balanced ImageNet-1K.%, as detailed in the appendix.
% Our method consistently delivers stronger performance under these extreme settings. Notably, under IF = 256 with ResNet as the verified model, our method achieves accuracy that not only outperforms all baselines in this highly imbalanced setting but also surpasses many existing methods trained on the full, balanced dataset (see Appendix for comparison).

\paragraph{Results under Extremely Low IPC Settings}
We further evaluate our method under severely compressed distillation regimes, where only one synthetic image per class is retained. 
% As shown in Table~\ref{ipc}, our method significantly outperforms all baselines, achieving over 20\% accuracy gain on CIFAR-10-LT and substantial improvements on CIFAR-100-LT and Tiny-ImageNet-LT. 
As shown in Table~\ref{ipc}, our method achieves over 2$\times$ accuracy improvement than all baselines on most datasets.
This strong performance stems from two key factors. First, fair BN statistic recalibration ensures that even a single synthetic image per class reflects accurate distribution-level information, providing reliable supervision despite minimal data capacity. Additionally, unbiased soft labels %from a teacher model 
provide semantic guidance that compensates for the limited representational expressiveness of low-IPC synthetic samples. Together, these mechanisms enable our method to remain robust under extreme distillation constraints.

% This strong performance is enabled by three factors specifically designed for low-IPC settings. First, our confidence-aware initialization maximizes information content per synthetic sample, ensuring each image is both class-representative and semantically aligned. Second, fair BN statistic recalibration ensures that the single image per class reflects accurate distribution-level information. Third, soft label relabeling with an unbiased teacher offers semantic guidance beyond the capacity of limited distilled samples.
% Together, these mechanisms enable effective supervision even when each class is represented by a single synthetic image, demonstrating the robustness of our approach under extreme distillation constraints.

% We further investigate the performance under extreme low-data regimes (IPC = 1), where learning meaningful representations is particularly challenging due to severe information bottlenecks. As shown in Table~\ref{ipc}, our method significantly outperforms all baselines, achieving over 20\% accuracy gain on CIFAR-10-LT and substantial margins on CIFAR-100-LT and Tiny-ImageNet-LT.
% This strong performance is not only attributed to the confidence-aware initialization strategy, which helps bootstrap informative features for each class, but also stems from the accurate estimation of class-wise BN statistics and the use of unbiased teacher-guided soft labels. Our approach enable the distilled data to reflect both structural regularity and semantic fidelity even when only a single instance per class is available.

\paragraph{Cross-Architecture Performance}
% To evaluate the generalizability of our distilled data across different network architectures, we conduct experiments using four commonly adopted backbones.
% To evaluate the generalizability of our distilled dataset across different evaluation architectures, we test its performance using four commonly adopted student model backbones.
% We assess the architecture-agnostic generalization of our distilled dataset by training different student models on the same synthetic data. 
To evaluate its generalization across architectures, we train multiple student models with different structures on the same distilled dataset.
As shown in Table~\ref{cross}, our method consistently outperforms existing approaches across four representative evaluation backbones.
Notably, baseline methods often show significant accuracy variation across architectures, while our distilled data supports uniformly strong performance. 
These results suggest that our method captures semantically meaningful and transferable patterns, facilitating generalizable learning across diverse student architectures.
%This confirms that our method distills semantically rich and transferable knowledge, enabling effective learning across diverse student networks.
%As shown in Table~\ref{cross}, our method consistently outperforms all prior approaches under each architecture. %, demonstrating strong cross-architecture transferability. 
% In particular, while baseline methods often exhibit large performance fluctuations depending on the chosen architecture, our approach maintains a high level of accuracy across both lightweight and deep models. This indicates that the synthetic data produced by our method captures intrinsic, architecture-agnostic semantic structures that can be effectively leveraged by a wide range of downstream models.

\paragraph{Class-wise Accuracy for Long-tailed Dataset}
Fig.~\ref{classwise} compares class-wise accuracy between DAMED~\cite{damed} and our method. DAMED underperforms on tail classes due to its biased expert training, which fails to preserve rare-class semantics. 
% Moreover, its gradient adjustment for trajectory  inadvertently suppressing mid-frequency class 
% performance. 
Moreover, the frequency-adjusted loss used during trajectory matching overlooks mid-frequency classes, resulting in suppressed performance.
% Moreover, the frequency-adjusted loss introduced to support trajectory matching inadvertently suppresses mid-frequency class performance.
In contrast, our method avoids these issues by first training a debiased expert model and then aligning fair BN statistics. 

%如表~\ref{ablation}所示，我们的每一个componen都对于long-tailed dataset distillation发挥着重要的作用。我们的model debiasing策略保留了更多tail类别的信息并使得recover过程也不会损失前面类别的性能，极大的提升了性能上限；statistics estimation模块使得尾部类别统计量的估计更加准确，能够充分发挥每一个样本的作用；adapted intialization策略为long-tail场景提供了diverse又有代表性的初始图象，即使在部分类别真实图像数量小于synthetic image的情况下也能够提供好的初始图像。这三部分共同使得uni-level dataset distillation framework能够配适long-tail setting.
\paragraph{Ablation on Different Components}
As shown in Table~\ref{ablation}, each component contributes critically to the success of our approach. The model debiasing strategy preserves tail-class semantics without degrading head or mid-frequency performance, thereby raising the overall performance ceiling. The recalibrated BN statistics ensure that each training sample, particularly those from low-shot categories, contributes fairly to the accumulated representation. The %adapted 
initialization strategy offers diverse, class-representative starting points for synthetic images, even when real data per class is scarce. 
% Together, these components form a cohesive framework well-suited to %the challenges of 
% long-tailed distributions. 
%To further verify the generality of our framework, we incorporate several state-of-the-art long-tailed classification methods into our pipeline in the appendix.
% each component in our framework plays a critical role in enhancing long-tailed dataset distillation. Our model debiasing strategy effectively preserves tail-class semantics while avoiding performance degradation on head or middle classes during the recovery process, thereby significantly improving the overall performance ceiling. The statistical recalibration module enables more accurate BN statistics, ensuring that each training sample—--especially in low-shot categories—-contributes fairly and effectively to supervision. The adapted initialization strategy provides diverse and representative starting points for synthetic images, even in cases where the number of real images per class is insufficient. These three components work in concert to make our uni-level dataset distillation framework well-adapted to the challenges of long-tailed distributions.

\paragraph{Computational Efficiency}
We assess the computational efficiency of our method by comparing both runtime and peak GPU memory usage with DAMED~\cite{damed}, the only existing approach explicitly designed for long-tailed dataset distillation.
% We assess computational and memory efficiency by comparing our method with DAMED~\cite{damed}, the only existing approach tailored for long-tailed dataset distillation.
As shown in Table~\ref{runtime}, our method substantially reduces the computation time required for both expert model training and distilled data synthesis. Specifically, across both datasets, the total runtime of our pipeline is less than one-twentieth that of DAMED.
In addition to faster execution, our method also exhibits more favorable memory behavior. 
% As shown in Table~\ref{memory}, the GPU memory usage of DAMED increases significantly with larger IPC values, which become a limiting factor in high-IPC scenarios.
As shown in Table~\ref{memory}, DAMED’s GPU memory usage grows rapidly with IPC, %severely 
constraining its applicability at higher values.
In contrast, our method maintains constant memory usage regardless of IPC, allowing for stable and efficient execution across a wide range of settings. %See Appendix for additional results and analysis.

% \begin{table}[htbp]
%     \centering
%     \small
%     \begin{tabular}{c|cc}
%     \hline Method & Expert Model Pretraining & Distilled Data Synthesis \\ 
%     \hline
%     \multicolumn{3}{c}{Dataset: CIFAR-10-LT (IF=100)} \\
%     \hline
%     DAMED   & 28511s + 2877s = 31388s& 30141s \\
%     Ours & & \\
%     \hline
%         \multicolumn{3}{c}{Dataset: CIFAR-100-LT (IF=50)} \\
%     \hline
%     DAMED   &21311s + 4958s = 26269s & 29328s \\
%     Ours & & \\
%     \hline
%     \end{tabular}
%     \caption{Runtime comparison on expert model pretraining and distilled data synthesis between DAMED~\cite{damed} and our approach.}
%     \label{runtime}
% \end{table}

\section{Conclusion}
We present a uni-level framework for long-tailed dataset distillation, explicitly designed to address the representation bias and inefficiency inherent in prior methods. We enhance distillation effectiveness under class imbalance via three key components: expert model debiasing, BN statistics recalibration, and confidence-aware initialization. Extensive experiments demonstrate that our method consistently outperforms existing baselines across a wide spectrum of IF and IPC settings, including highly imbalanced and low-sample regimes, showcasing strong robustness and generalizability. 
%Unlike previous approaches that either assume balanced data or suffer from high computational costs, 
% We systematically improve distillation effectiveness in imbalanced settings through three key components: model-level debiasing for both observer and teacher networks, fair recalibration of BN statistics, and a tailored initialization strategy for long-tailed settings. Extensive experiments on diverse benchmarks demonstrate that our method consistently outperforms existing baselines under various IF and IPC settings, achieving substantial gains in both accuracy and computational efficiency.
\paragraph{Broader Impact}
Our method can potentially be extended to multi-domain or federated dataset distillation, where data imbalance naturally occurs across clients or domains. %Moreover, the proposed strategies may benefit other tasks requiring compact yet fair data summarization, such as continual learning or low-shot model pretraining.

\section{Acknowledgments}
This work was supported by the GPU cluster of the MCC Lab, Information Science and Technology Institution, and the Supercomputing Center of USTC.
%This work was supported by the GPU cluster built by MCC Lab of Information Science and Technology Institution, USTC, and the Supercomputing Center of the USTC.

%\newpage
\bibliography{aaai2026}

\end{document}